\documentclass[10pt,twocolumn,letterpaper]{article}

\usepackage[pagenumbers]{cvpr} % To force page numbers, e.g. for an arXiv version
\usepackage{algorithm}
\usepackage{algorithmic}
\usepackage{bm}
\usepackage{marvosym}  % 提供 \Letter 命令

\definecolor{cvprblue}{rgb}{0.21,0.49,0.74}
\usepackage[pagebackref,breaklinks,colorlinks,allcolors=cvprblue]{hyperref}

\def\paperID{} % *** Enter the Paper ID here
\def\confName{CVPR}
\def\confYear{2026}

\title{Just-in-Time: Training-Free Spatial Acceleration for Diffusion Transformers}

\author{%
Wenhao Sun$^{1}$ \quad
Ji Li$^{2}$ \quad
Zhaoqiang Liu$^{1,\text{\Letter}}$
\\[0em]
$^{1}$University of Electronic Science and Technology of China \quad
$^{2}$Capital Normal University \quad \\
\\[-1.2em]
{\small \texttt{202511081644@std.uestc.edu.cn} \quad \texttt{matliji@163.com} \quad \texttt{zqliu12@gmail.com}} \\
{Project Page: \href{https://wenhao-sun77.github.io/JiT/}{https://wenhao-sun77.github.io/JiT/}} \\
}

\begin{document}
\twocolumn[{
    \renewcommand\twocolumn[1][]{#1}
    \maketitle
    \begin{center}
    \vspace{-25px}
    \includegraphics[width=0.9\linewidth]{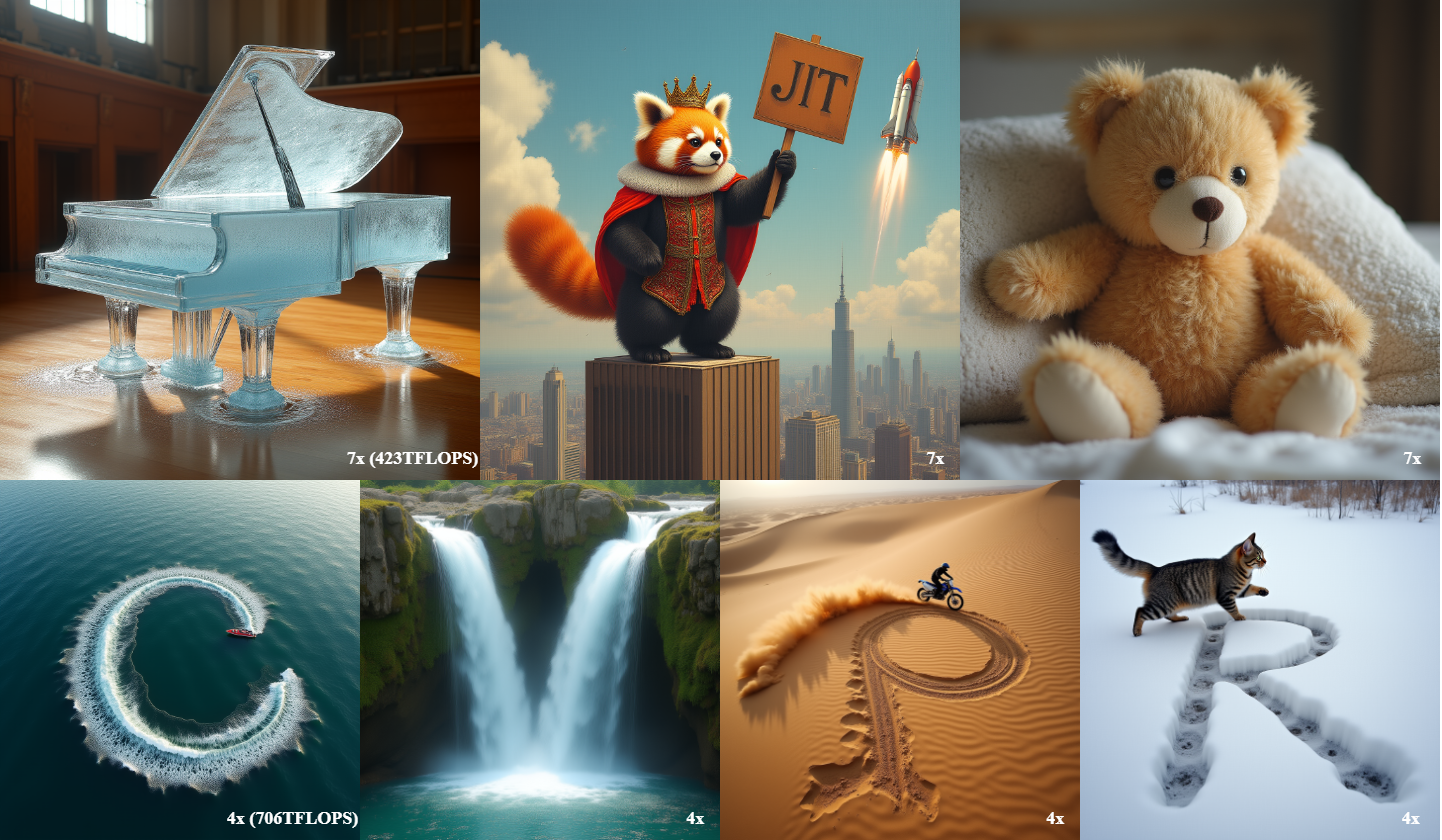}
    \captionof{figure}{Visual showcases of our JiT framework applied to the FLUX.1-dev model. Our method produces high-fidelity and visually compelling images even at significant acceleration factors of \(4\times\) and \(7\times\).}
    \label{fig:kt}
    \end{center}
}]

\maketitle

\begingroup
\renewcommand{\thefootnote}{\fnsymbol{footnote}}
\footnotetext{
  $^{\text{\Letter}}$Corresponding author. \quad
}
\endgroup

\begin{abstract}
Diffusion Transformers have established a new state-of-the-art in image synthesis, but the high computational cost of iterative sampling severely hampers their practical deployment. While existing acceleration methods often focus on the temporal domain, they overlook the substantial spatial redundancy inherent in the generative process, where global structures emerge long before fine-grained details are formed. The uniform computational treatment of all spatial regions represents a critical inefficiency. In this paper, we introduce \textbf{Just-in-Time (JiT)}, a novel training-free framework that addresses this challenge by acceleration in the spatial domain. JiT formulates a spatially approximated generative ordinary differential equation (ODE) that drives the full latent state evolution based on computations from a dynamically selected, sparse subset of anchor tokens. To ensure seamless transitions as new tokens are incorporated to expand the dimensions of the latent state, we propose a deterministic micro-flow, a simple and effective finite-time ODE that maintains both structural coherence and statistical correctness. Extensive experiments on the state-of-the-art FLUX.1-dev model demonstrate that JiT achieves up to a \textbf{7\(\times\) speedup} with nearly lossless performance, significantly outperforming existing acceleration methods and establishing a new and superior trade-off between inference speed and generation fidelity.
\end{abstract}   
\vspace{-10px}
\section{Introduction}
\label{sec:intro}
Generative artificial intelligence is fundamentally reshaping the paradigm of digital content creation at an unprecedented depth and breadth.
Among its advancements, text-to-image~\cite{sd,t2i1,t2i2,t2i3} and text-to-video~\cite{video1,video2,video3,video4,video5} synthesis, underpinned by diffusion models~\cite{ddpm, score,dm1}, have become the dominant technological route due to their exceptional generation quality and semantic controllability.
In recent developments, the diffusion Transformer (DiT)~\cite{dit,dit1,dit2} has emerged as a particularly potent architecture.
By leveraging self-attention mechanisms~\cite{attention} to efficiently model long-range dependencies between arbitrary tokens in image or video data, DiT have demonstrated remarkable scalability and performance potential, continuously pushing the frontiers of generative fidelity.

However, the exceptional performance of DiT is coupled with a substantial computational overhead stemming from its self-attention mechanism, whose complexity grows quadratically with the number of input tokens.
When generating high-resolution images or, more critically, long-duration videos, this growth leads to a computational avalanche.
This issue is further exacerbated by the inherently iterative nature of the diffusion denoising process, resulting in prohibitive inference latency and demanding stringent hardware requirements.
These challenges severely constrain the application of DiT in real-time interactive systems, consumer-grade devices, and large-scale commercial services.

Efforts to mitigate the high computational cost of DiT have largely progressed along two main directions. The first focuses on the temporal domain, striving to shorten the sampling trajectory by reducing the number of function evaluations. This line of work includes the design of sophisticated high-order solvers~\cite{slover1,slover2,zhao2023unipc,xue2024accelerating,ma2024surprising} and the distillation of pre-trained models into few-step variants~\cite{dis1,dis2,di3,di4,fu2025learnable}. However, these approaches often come with the caveats of potentially compromising generation fidelity in ultra-low step regimes or demanding retraining resources. A second (complementary) direction aims to make each sampling step more efficient. This is achieved through techniques such as feature caching, which exploits temporal redundancy by reusing intermediate activations across iterations~\cite{cache1, cache2,cache3,cache4}, and low-bit quantization, which accelerates underlying matrix operations and reduces the memory footprint by representing weights and activations with lower numerical precision~\cite{qu1, qu2,qu3}.

Despite their notable successes in optimizing temporal steps and model parameters, these approaches often overlook the intrinsic spatial redundancy inherent in the generative process.
Specifically, they treat all spatial regions with uniform computational effort throughout the generation process.
We consider this to be an unnecessary luxury, given the well-established characteristic of diffusion models to first synthesize low-frequency global structures before progressively refining high-frequency details.
This observation raises a natural question: How can we dynamically focus computational resources on salient spatial regions during the early stages of generation while deferring computation on less critical areas?

To realize this opportunity for dynamic spatial acceleration, we introduce Just-in-Time (JiT), a novel and training-free framework for spatial acceleration. Specifically, our approach achieves dynamic spatial acceleration by introducing a reasonable approximation to the underlying generative ordinary differential equation (ODE), which enables a dynamic reduction of computational load in the spatial domain without requiring model retraining. We implement JiT on the state-of-the-art (SOTA) DiT model FLUX.1-dev~\cite{flux} and demonstrate that it attains nearly lossless performance while achieving significant inference speedups of up to \(7\times\) (as shown in Fig.~\ref{fig:kt}). Our main contributions are summarized as follows:
\begin{itemize}
    \item We propose JiT, a novel and training-free framework for accelerating spatial image generation in flow-matching-based DiT models. Our framework is meticulously designed with two synergistic components: (1) The spatially approximated generative ODE (SAG-ODE), which efficiently drives the latent state by approximating the full velocity field from sparse anchor tokens computations using an augmented lifter operator. (2) The deterministic micro-flow (DMF), a well-designed finite-time ODE that ensures seamless and statistically correct activation of new tokens during stage transitions, thereby maintaining latent space continuity and preventing artifacts.
    \item Extensive experiments and user studies demonstrate the superior acceleration capabilities and preserved generation quality of our method, outperforming existing SOTA acceleration methods for image generation.
\end{itemize}

\vspace{-7px}
\section{Related works}
\label{sec:related works}
\subsection{Diffusion models}
Denoising diffusion probabilistic models (DDPMs)~\cite{ddpm} have become a dominant force in generative modeling, lauded for their high-fidelity synthesis capabilities. Their formulation has evolved from score-based stochastic differential equations (SDEs)~\cite{score} to deterministic ODE samplers, such as DDIM~\cite{ddim}, and more recently to the flow matching paradigm~\cite{flow1,flow2}, which directly learns the ODE vector field and is a framework exemplified by FLUX.1-dev model used in our work. Despite their prowess, a critical limitation remains in their slow, iterative sampling process, requiring a large number of neural function evaluations (NFEs). This computational burden has spurred the development of various acceleration methods, and our work contributes a novel, training-free spatial acceleration technique tailored for this modern ODE-based framework.
\vspace{-5px}
\subsection{Spatial acceleration}
Spatial acceleration methods aim to reduce the substantial computational cost of diffusion models by operating on a spatially reduced representation of the latent state, particularly during the early, noise-dominated stages of the generation process. Early approaches~\cite{ralu,bottle,jen} adopted a pyramidal or hierarchical strategy~\cite{pyramidal,cascaded,dimensionality}, generating a low-resolution latent and progressively upscaling it. These methods typically rely on explicit upsampling or downsampling operators for dimensionality transition, often coupled with a distribution correction method to align the rescaled features with the target noise distribution at each new scale.

While these hierarchical methods have demonstrated feasibility, they are often plagued by some major challenges. For instance, upsampling operators can introduce information loss or aliasing artifacts, and the efficacy of post-hoc correction steps is not always guaranteed, frequently leading to visual inconsistencies in the final output. A notable departure from this paradigm is subspace diffusion~\cite{jing2022subspace}, which showed that generation could be confined to low-dimensional subspaces without explicit resizing. Drawing inspiration from this concept, we introduce the first training-free method to completely obviate the need for such error-prone upsampling and correction procedures, directly manipulating token subspaces to achieve seamless and artifact-free dimensionality transitions.

\vspace{-7px}
\section{Methodology}
\begin{figure*}[t]
  \centering

   \includegraphics[width=0.87\linewidth]{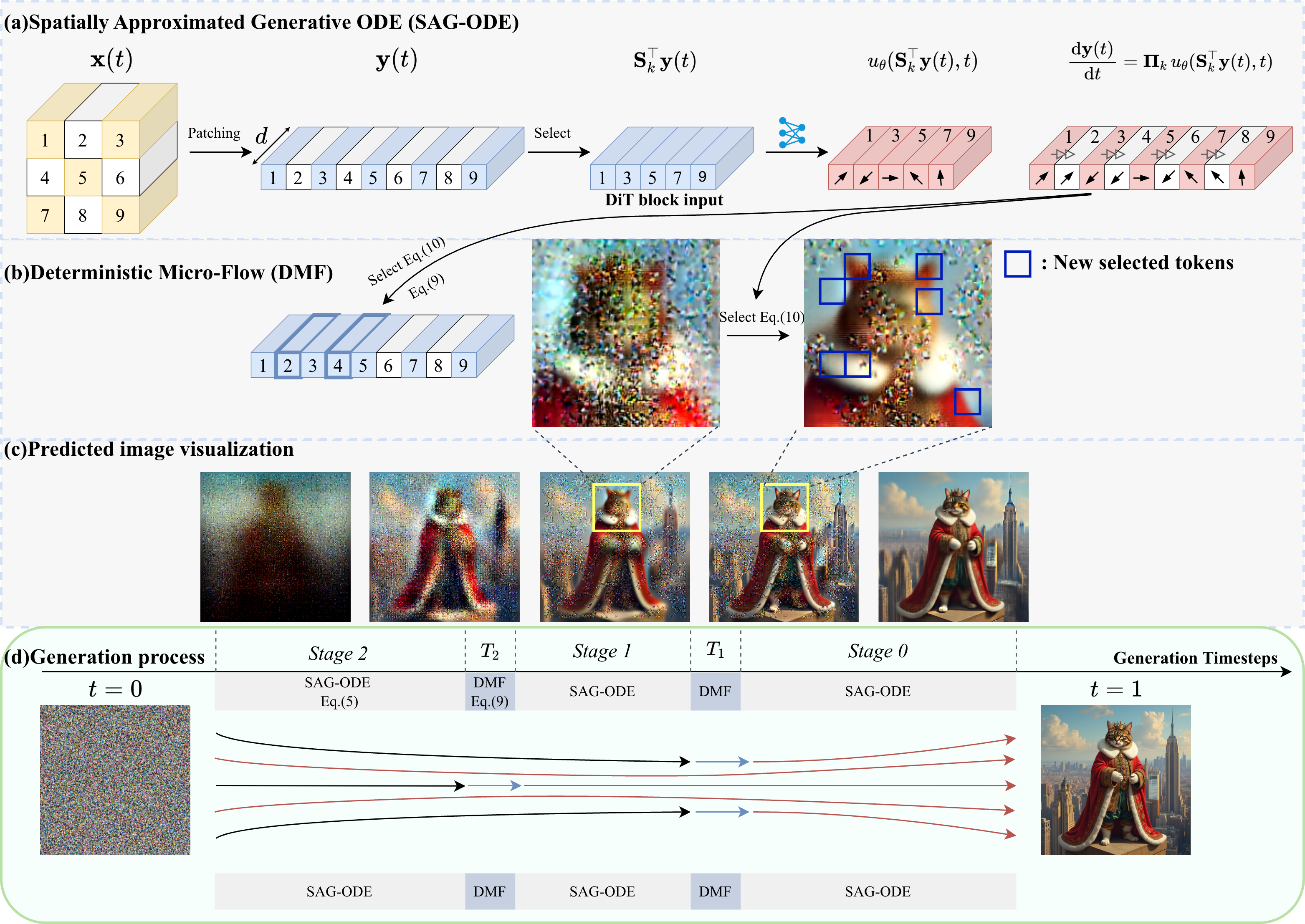}

  \caption{An overview of our JiT framework, illustrating its core mechanisms and underlying philosophy. 
\textbf{(a)} The SAG-ODE evolves the latent state by extrapolating a velocity field computed on a sparse subset of tokens. 
\textbf{(b)} For stage transitions, the DMF evolves newly incorporated tokens to a structurally coherent target with the correct noise level to prevent artifacts. 
\textbf{(c)} The visualized evolution of the predicted clean image reveals a coarse-to-fine process (global structures first), motivating our strategy to defer computation on detailed regions.
\textbf{(d)} The sampling trajectory visualizes our dynamic resource allocation, where the set of active tokens (red flow) starts as a narrow subset and expands over time, reserving full computation for the final detail-refining stages.}
   \label{fig:overview}
\end{figure*}

The overall framework diagram of the proposed method is depicted in Fig.~\ref{fig:overview}. There are two principal components, namely the spatially approximated generative ODE (SAG-ODE) and deterministic micro-flow (DMF), which will be elucidated in more detail in the following subsections.

\subsection{Preliminaries}

\paragraph{Flow matching and tokenized generation}
Flow Matching formulates generative modeling as solving an ODE that establishes a deterministic transport between a noise distribution $p_0$ and the data distribution $p_1$.
This process typically operates on a compressed latent representation, obtained by encoding an image through a variational autoencoder (VAE)~\cite{vae}. This yields a latent tensor \(\mathbf{x} \in \mathbb{R}^{C \times H \times W}\), where \(C\) denotes the number of channels, and \(H\) and \(W\) are its spatial height and width. The flow is then defined over this continuous latent space from noise \(\mathbf{x}(0)\) to data latents \(\mathbf{x}(1)\).

For practical implementation with DiT, the continuous latent representation $\mathbf{x}(t)$ is first tokenized via spatial patching into a sequence of $N$ tokens, which we denote as $\mathbf{y}(t) \in \mathbb{R}^{N d}$, where $d$ is the dimension of the token.
The Transformer network $\bm{u}_{\bm{\theta}}$ takes this token sequence as input to model the underlying velocity field. The ODE thus governs the dynamics of the tokenized representation:
\begin{equation}
\frac{\mathrm{d}\mathbf{y}(t)}{\mathrm{d}t} = \bm{u}_{\bm{\theta}}(\mathbf{y}(t), t), \quad t \in [0, 1],
\label{eq:rectified_flow}
\end{equation}
where $t=0$ corresponds to tokens of pure Gaussian noise and $t=1$ corresponds to tokens of the data latent.
During inference, we solve this ODE from $t=0$ to $t=1$ to generate data tokens, which are then de-tokenized and decoded by the VAE to produce the final image.
The primary computational bottleneck arises here, as $\bm{u}_{\bm{\theta}}$ must process all $N$ tokens at each timestep, incurring an $\mathcal{O}(N^2)$ cost.

\paragraph{Nested token-subset chain and projectors}
To address this bottleneck, our method operates on a dynamically changing subset of tokens that are fed into the Transformer for computation. We refer to these tokens as \textit{activated anchor tokens}. Formally, we construct a nested hierarchy of index sets for these anchor tokens that represents a coarse-to-fine generation strategy:
\begin{equation}
\Omega_K \subset \Omega_{K-1} \subset \cdots \subset \Omega_1 \subset \Omega_0 = \{1, 2, \ldots, N\},
\end{equation}
where \(|\Omega_k| = m_k\) is the number of anchor tokens at stage $k$. Meanwhile, we refer to tokens not included in the current token set $\Omega_k$ as inactive tokens.
For each subset $\Omega_k$, we define a selector matrix $\mathbf{S}_k \in \{0,1\}^{Nd \times m_kd}$ that extracts the corresponding anchor tokens $\mathbf{y}_k = \mathbf{S}_k^\top \mathbf{y}$ from the full token sequence.
Consequently, the orthogonal projector onto the subspace spanned by the anchor tokens is given by the matrix \(\mathbf{P}_k \in \mathbb{R}^{Nd \times Nd}\), defined as:
\begin{equation}
\mathbf{P}_k :=  \mathbf{S}_k\mathbf{S}_k^\top.
\end{equation}
This projector acts as a mask on the token dimension. Furthermore, we define a projector \(\mathbf{Q}_k \in \mathbb{R}^{Nd \times Nd}\) onto the subspace of newly activated tokens, which is given by the difference between two consecutive projectors:
\begin{equation}
\mathbf{Q}_k := \mathbf{P}_{k-1} - \mathbf{P}_k.
\label{eq:ring_projector}
\end{equation}
This operator precisely isolates the set of tokens \(R_k:= \Omega_{k-1} \setminus \Omega_k\), which are to be activated at the next stage transition. These operators are fundamental building blocks for constructing our accelerated generative ODE in the following section.

\vspace{-5px}

\subsection{The spatially approximated generative ODE}
At the heart of our framework lies a simplified and elegant generative ODE that governs the evolution of the tokenized latent state \(\mathbf{y}(t)\) during each stage of the sampling process. We refer to it as SAG-ODE. For a given stage \(k\) operating on the activated anchor token set \(\Omega_{k}\), the SAG-ODE is described by:
\begin{equation}
\frac{\mathrm{d}\mathbf{y}(t)}{\mathrm{d}t} = \mathbf{\Pi}_{k}\, \bm{u}_{\bm{\theta}}(\mathbf{S}^\top_{k} \mathbf{y}(t), t) = \bm{v}_t.
\label{eq:main_ode_simplified}
\end{equation}
The efficiency of our SAG-ODE is enabled by the augmented lifter operator, \(\mathbf{\Pi}_{k}: \mathbb{R}^{m_k d} \to \mathbb{R}^{N d}\). It takes the velocity field \(\bm{u}_{\bm{\theta}}(\mathbf{S}^\top_{k} \mathbf{y}(t), t)\in \mathbb{R}^{m_k d}\), computed by the Transformer on only the \(m_{k}\) anchor tokens, and extrapolates it to define a complete, full-space velocity field that drives the dynamics of all \(N\) tokens.

The lifter \(\mathbf{\Pi}_{k}\) constructs this full velocity field by performing two concurrent functions:
\begin{equation}
\mathbf{\Pi}_k \bm{u}_{\bm{\theta}} := \mathbf{S}_k \bm{u}_{\bm{\theta}} + \mathcal{I}_k(\bm{u}_{\bm{\theta}}).
\label{eq:lifter}
\end{equation}
Here, the first term \(\mathbf{S}_k \bm{u}_{\bm{\theta}}\) acts as an embedding map, precisely placing the computed velocity \(\bm{u}_{\bm{\theta}}\) back into its corresponding positions in full-space for the anchor tokens. This forms the exact part of the ODE dynamics. Concurrently, the interpolation operator $\mathcal{I}_k$ (detailed in Appendix~\ref{inter}) efficiently approximates the velocity for the inactive subspace via smooth spatial interpolation from \(\bm{u}_{\bm{\theta}}\). In doing so, the single operator \(\mathbf{\Pi}_{k}\) cohesively defines a high-quality, structurally-aware velocity field that drives the SAG-ODE, guided solely by the computations on the anchor set. A simplified schematic diagram of the SAG-ODE is shown in Fig.~\ref{fig:overview}(a).

An important feature of our SAG-ODE is its \textit{consistency property}, which ensures that the approximation introduces zero error on the anchor tokens themselves. By design, the interpolation operator \(\mathcal{I}_k\) does not affect the anchor token subspace, i.e., \(\mathbf{S}^\top_k \mathcal{I}_k(\bm{u}_{\bm{\theta}}) = \mathbf{0}\). This leads to the property:
\begin{equation}
\mathbf{S}^\top_k \left( \mathbf{\Pi}_k \bm{u}_{\bm{\theta}} \right) = \mathbf{S}^\top_k (\mathbf{S}_k \bm{u}_{\bm{\theta}} + \mathcal{I}_k(\bm{u}_{\bm{\theta}})) = \bm{u}_{\bm{\theta}}.
\label{eq:consistency}
\end{equation}
Applying this to our SAG-ODE (see Eq.~\eqref{eq:main_ode_simplified}) confirms that the dynamics of the anchor tokens are governed exactly by the output of the Transformer with \(\mathbf{S}^\top_k \frac{\mathrm{d}\mathbf{y}(t)}{\mathrm{d}t} = \bm{u}_{\bm{\theta}}(\mathbf{S}^\top_k \mathbf{y}(t), t)\). This property guarantees that our acceleration strategy does not compromise the learned dynamics on the computationally critical parts of the state. Furthermore, it ensures that these anchor tokens are continuously updated within the full-dimensional context, allowing them to maintain their correct statistical properties throughout the entire generation process. A formal derivation is provided in Appendix~\ref{A}.

\vspace{-5px}

\subsection{Deterministic micro-flow}
\label{sec:variance}
While the SAG-ODE in Eq.~\eqref{eq:main_ode_simplified} governs the evolution for a fixed set of anchor tokens $\Omega_k$, a coherent mechanism is required to handle the transition between stages, specifically the activation of new tokens to expand the subspace dimension. A naive or instantaneous injection of state for these new tokens risks creating spatial discontinuities and statistical mismatches, which can propagate as artifacts in the final output. To ensure a seamless and theoretically grounded transition, we propose the DMF. This short, finite-time procedure deterministically evolves the state of newly activated tokens from their interpolated state to a statistically correct target.

At a transition time \(T_k\), we activate new tokens corresponding to the index set \(R_k\). The first step is to construct a sound target state, \(\mathbf{y}_k^\star\), for these new tokens that fuses structural information from existing anchor tokens with the correct noise level. We define the target as
\begin{equation}
\mathbf{y}_k^\star = \mathbf{Q}_k \left( T_k \Phi_k(\mathbf{S}^\top _k \mathbf{\hat{y}}(1)) + (1-T_k)\epsilon \right),
\label{eq-target-state}
\end{equation}
where the predicted data latent tokens $\mathbf{\hat{y}}(1)=\mathbf{y}_{t_{i-1}}+(1-t_{i-1})\bm{v}_{t_{i-1}}$ is derived from Tweedie's formula~\cite{lipman2024flow} evaluated at the preceding timestep $t_{i-1}$. Here, $\Phi_k$ denotes the unified structural prior operator that spatially interpolates this clean prediction to the full spatial dimension (detailed in Appendix~\ref{inter}), and $\epsilon \sim \mathcal{N}(\bm{0}, \bm{I})$ is a standard full-dimensional Gaussian noise. (The visualization of the predicted data is shown in Fig.~\ref{fig:overview}(c).) This formulation ensures that the newly activated tokens are not only structurally coherent but also statistically consistent with the flow matching trajectory, thereby preventing distribution shifts and providing a stable transition.

With the target state established, the evolution of the new tokens \(\mathbf{Q}_k \mathbf{y}\) is governed by a finite-time hitting ODE over a short interval \([T_k - \delta, T_k]\), where \(\delta\) is an extremely small duration. This ODE is defined as:
\begin{equation}
\mathbf{Q}_k \dot{\mathbf{y}}(t) = \frac{\mathbf{y}_k^\star - \mathbf{Q}_k \mathbf{y}(t)}{T_k - t}, \quad t \in [T_k - \delta, T_k].
\label{eq:microflow_ode}
\end{equation}
The time-varying rate \((T_k - t)^{-1}\) ensures that the state \(\mathbf{Q}_k \mathbf{y}(t)\) is driven from its initial value at \(t=T_k-\delta\) and precisely converges to the target \(\mathbf{y}_k^\star\) at the end of the interval, i.e., \(\lim_{t \to T_k} \mathbf{Q}_k \mathbf{y}(t) = \mathbf{y}_k^\star\). During this micro-flow, the velocity of existing anchor tokens is set to zero (i.e., \(\mathbf{P}_k \dot{\mathbf{y}}(t) = \mathbf{0}\)) to isolate the transition dynamics to the new tokens. This procedure guarantees a smooth, continuous trajectory in the latent space, which is effective for preventing artifacts and maintaining the stability of the overall generation process.

\paragraph{Importance-guided token activation}
\label{sec:token_selection}
Instead of activating new tokens based on a predefined, static pattern, we introduce a dynamic selection strategy importance-guided token activation (ITA) that prioritizes regions of high information density. As illustrated in Fig.~\ref{fig:overview}(b), once the low-frequency structure of an image is established, subsequent computational effort should be directed towards synthesizing high-frequency details, such as contours and textures. Our intuition is that these details emerge in regions where the generative process is most active, corresponding to areas where the velocity field \(\bm{u}_{\bm{\theta}}(\mathbf{y}(t), t)\) exhibits the most significant local variation. We therefore quantify this dynamic activity by measuring the local variance of the velocity field predicted by the DiT.

Specifically, at each transition time \(T_k\), we compute a spatial importance map \(\textbf{I}(T_k)\). The importance score at each token location is defined as the variance of the velocity vectors within a small local patch (e.g., \(3 \times 3\)). This local variance can be computed efficiently using average pooling operations on both the velocity field and its element-wise square. Specifically, letting \(\mathcal{W}\) denote the local averaging window, the importance map is computed as
\begin{align}
&\textbf{I}(t) = \mathbb{E}_{\mathcal{W}}[\bm{u}_{\bm{\theta}}(\mathbf{y}(t), t) \odot \bm{u}_{\bm{\theta}}(\mathbf{y}(t), t)] \nonumber \\
& \quad - \left( \mathbb{E}_{\mathcal{W}}[\bm{u}_{\bm{\theta}}(\mathbf{y}(t), t)] \right) \odot \left( \mathbb{E}_{\mathcal{W}}[\bm{u}_{\bm{\theta}}(\mathbf{y}(t), t)] \right),
\label{eq:variance_map}
\end{align}
where \(\odot\) denotes the element-wise product. We then rank all inactive tokens based on their corresponding importance scores and select the top candidates to form the newly activated token set \(R_k\). As shown in Fig.~\ref{fig:overview}(b), this content-aware strategy ensures that computational resources are allocated to the most dynamic spatial regions, leading to a more efficient and effective synthesis of fine-grained details.

\begin{algorithm}[ht]
\caption{The JiT Sampling Algorithm}
\label{alg:s3_amsrf_formal}
\begin{algorithmic}[1]
\STATE \textbf{Input:} Pre-trained DiT $\bm{u}_{\bm{\theta}}$, total generation steps $N$, schedule $\{T_k, m_k\}_{k=0}^K$, time schedule $\{t_i\}_{i=0}^N$, VAE decoder $\mathrm{D}_{\bm{\phi}}$, Gaussian noise $\epsilon \sim \mathcal{N}(\bm{0}, \bm{I})$

\STATE \textbf{Output:} Generated image $\mathbf{x}$

\STATE Initialize full-dimensional state vector $\mathbf{y}_{t_0} \sim \mathcal{N}(\bm{0},\bm{I})$
\STATE Initialize the smallest token set $\Omega_K$ based on $m_K$
\STATE Set current stage $k \leftarrow K$
\FOR{$i = 0, \dots, N-1$}
  
    \IF{$t_{i} == T_k$ \textbf{and} $k > 0$}
        \STATE \COMMENT{\textit{--- Stage Transition ---}}
    \STATE Compute importance map $\textbf{I}(T_k)$ :
    \STATE $\textbf{I}(T_k) \leftarrow \mathbb{E}_{\mathcal{W}}[\bm{v}_{t_{i-1}} \odot \bm{v}_{t_{i-1}} ] - \left( \mathbb{E}_{\mathcal{W}}[\bm{v}_{t_{i-1}}] \right) \odot \left( \mathbb{E}_{\mathcal{W}}[\bm{v}_{t_{i-1}}] \right)$
    \STATE Select new tokens to activate: 
    \STATE $R_k \leftarrow \text{TopTokens}(\textbf{I}(T_k),m_{k-1} - m_k)$
    \STATE $\Omega_{k-1} \leftarrow \Omega_k \cup R_k$
    \STATE Predicted data latent tokens $\mathbf{\hat{y}}(1)$:
    \STATE $\mathbf{\hat{y}}(1) \leftarrow \mathbf{y}_{t_{i-1}} + (1-t_{i-1})\bm{v}_{t_{i-1}}$ 
    
    \STATE Set target state $\mathbf{y}_k^\star$ for the newly activated tokens:
    \STATE $\mathbf{y}_k^\star = \mathbf{Q}_k \left( T_k \Phi_k(\mathbf{S}^\top _k \mathbf{\hat{y}}(1)) + (1-T_k)\epsilon \right)$
    \STATE $\mathbf{Q}_k \mathbf{y}_{t_i} \leftarrow \mathbf{y}_k^\star$
    \STATE $k \leftarrow k - 1$
  \ENDIF
  \STATE Set step size: $\Delta t \leftarrow t_{i+1} - t_i$
  \STATE Compute full-dimensional velocity $\bm{v}_{t_i}$:
  \STATE $\bm{v}_{t_i} \leftarrow \mathbf{\Pi}_{k}\, \bm{u}_{\bm{\theta}}(\mathbf{S}^\top_{k} \mathbf{y}_{t_i}, t_i)$
  \STATE $\mathbf{y}_{t_{i+1}} \leftarrow \mathbf{y}_{t_i} + \bm{v}_{t_i}\cdot \Delta t$
\ENDFOR
\STATE $\mathbf{x} \leftarrow \mathrm{D}_{\bm{\phi}}(\mathbf{y}_{t_{N}})$
\RETURN $\mathbf{x}$
\end{algorithmic}
\end{algorithm}

\vspace{-5px}

\subsection{The JiT sampling algorithm}
The complete JiT sampling process is summarized in Algorithm~\ref{alg:s3_amsrf_formal}.

\vspace{-7px}
\section{Experiments}
\subsection{Experimental setup}
\paragraph{Baselines}
We evaluate our method JiT on FLUX.1-dev \cite{flux}, a leading DiT trained with the flow matching framework. Our approach is benchmarked against several SOTA acceleration baselines, which include two spatial-domain methods, namely RALU~\cite{ralu} and Bottleneck Sampling~\cite{bottle}, and two caching-based methods, namely TaylorSeer~\cite{cache2} and TeaCache~\cite{cache1}. For a comprehensive comparison, we also include the performance of the vanilla FLUX.1-dev pipeline executed with varying NFEs. We measured acceleration performance using latency and FLOPs on a single A800 GPU. Appendix \ref{sec:appendix_b} provides further implementation details for JiT.

\paragraph{Evaluation metrics}
We evaluate performance from two key perspectives: image generation quality and text-image alignment. For generation quality, we employ the no-reference metric CLIP-IQA \cite{clipiqa} alongside two leading human preference models, ImageReward \cite{imagereward} and HPSv2.1 \cite{hps}. For text-image alignment, we utilize the GenEval \cite{geneval} and T2I-CompBench \cite{t2icompbench} benchmarks for evaluation.

\begin{table*}[t]
  \renewcommand{\arraystretch}{0.95} 
  \centering
  \resizebox{\textwidth}{!}{%
  \begin{tabular}{lccccccccc}
    \toprule
    \textbf{Method} & \textbf{NFE} & \textbf{Latency (s)} $\downarrow$ & \textbf{TFLOPs} $\downarrow$ & \textbf{Speed} $\uparrow$ & \textbf{CLIP-IQA} $\uparrow$ & \textbf{Image Reward} $\uparrow$ & \textbf{HPSv2.1} $\uparrow$ & \textbf{Gen Eval} $\uparrow$ & \textbf{T2I-Comp.} $\uparrow$ \\
    \midrule
    FLUX.1-dev & 50 & 25.25 & 2990.96 & 1.00$\times$ & 0.6139 & 1.004 & 30.39 & 0.6565 &  0.4836\\
    \midrule
    FLUX.1-dev & 12 & 6.21 & 729.07 & 4.10$\times$ & 0.5341 & 0.9435 & 29.43 & 0.6426 &  0.4843\\
    Bottleneck & 14 & 6.37 & 709.28 & 4.22$\times$ & 0.4056 & 0.9216 & 29.01 & 0.6499 &  0.4779\\
    RALU       & 18 & 6.23 & 723.69 & 4.13$\times$ & 0.5493 & 0.8560 & \textbf{29.83} & 0.6328 &  0.4620\\
    Teacache   & 28 & 6.98 & 729.32 & 4.10$\times$ & 0.6003 & 0.9638 & 29.68 & 0.6493 &  0.4849\\
    \textbf{JiT (Ours)} & 18 & \textbf{6.02} & \textbf{706.17} & \textbf{4.24$\times$} & \textbf{0.6166} & \textbf{1.017} & \underline{29.77} & \textbf{0.6540} &  \textbf{0.4991}\\
    \midrule
    FLUX.1-dev & 7  & 3.80 & 431.44 & 6.93$\times$ & 0.4134 & 0.7474 & 27.72 & 0.5629 & 0.4635\\
    RALU       & 10 & 3.85 & 426.01 & 7.02$\times$ & 0.4865 & 0.8904 & \textbf{29.71} & 0.6394 &  0.4731\\
    Taylorseer & 28 & 5.20 & 432.09 & 6.92$\times$ & 0.4164 & 0.7995 & 28.46 & 0.5832 &  0.4583\\
    Teacache   & 28 & 4.53 & 431.79 & 6.92$\times$ & 0.5183 & 0.7731 & 27.86 & 0.5837 &  0.4625\\
    \textbf{JiT (Ours)} & 11 & \textbf{3.67} & \textbf{423.26} & \textbf{7.07$\times$} & \textbf{0.5397} & \textbf{0.9746} & \underline{29.02} & \textbf{0.6457} &  \textbf{0.4961}\\
    \bottomrule
  \end{tabular}%
  }
  \caption{Quantitative comparison with other methods. The optimal result is represented in bold, and the sub-optimal result is represented in underlining. FLOPs are measured via \texttt{torch.profiler} and the Speed is calculated based on the FLOPs reduction relative to the base model FLUX.1-dev(50).}
  \label{tab:Quantitative}
\end{table*}

\vspace{-5px}

\subsection{Quantitative results}
As detailed in \cref{tab:Quantitative}, our JiT framework achieves SOTA performance at both \(\sim\)4\(\times\) and \(\sim\)7\(\times\) acceleration tiers, maintaining nearly lossless quality compared to the high-fidelity 50-NFE baseline. This contrasts sharply with competing approaches, where cache-based methods remain fundamentally capped by the quality of the low-NFE baseline, and alternative spatial methods introduce artifacts through explicit upsampling and correction steps. In contrast, JiT avoids these inherent limitations. Its upsampling-free design leverages the intrinsic multi-scale knowledge of the DiT, while the seamless DMF ensures artifact-free state transitions. Consequently, JiT establishes a new, superior speed-fidelity trade-off, maintaining high quality even at a demanding \(7\times\) speedup.

\subsection{Qualitative results}
\begin{figure*}[ht]
  \centering

   \includegraphics[width=0.95\linewidth]{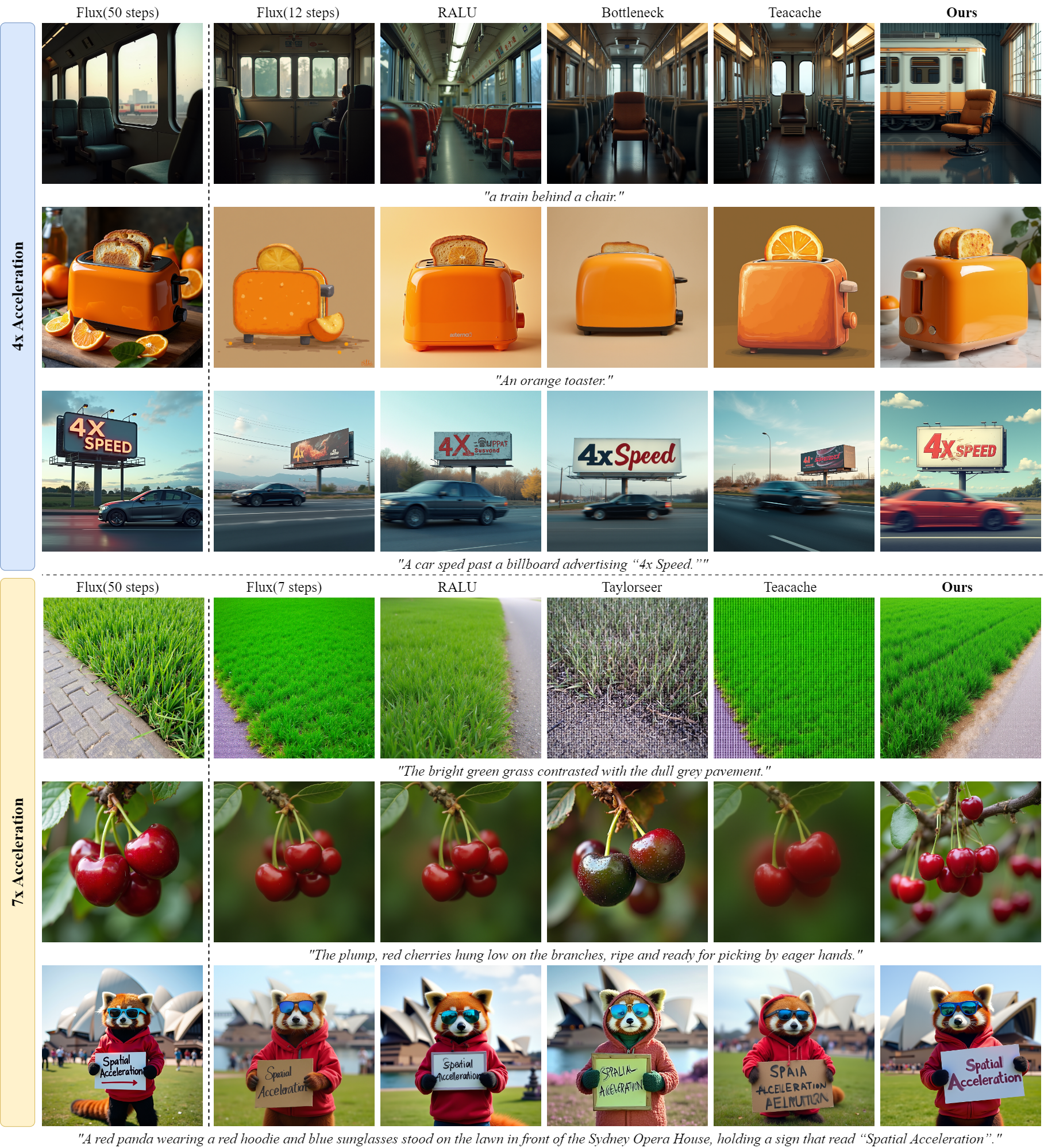}

   \caption{Qualitative comparison demonstrating the superior performance of our JiT framework. While competing methods suffer from common acceleration artifacts, including semantic errors, loss of fine detail, and structural inconsistencies, our approach maintains high fidelity across different prompts and acceleration levels.}
   \label{fig:Qualitative}
\end{figure*}

For a qualitative assessment of generation fidelity, we present a series of visual comparisons in Fig.~\ref{fig:Qualitative}. These examples provide tangible evidence for the quantitative findings in Tab.~\ref{tab:Quantitative} and highlight the distinct advantages of our JiT framework, especially under \(7\times\) acceleration settings.

As shown, competing acceleration methods exhibit noticeable degradation in quality. Several baselines struggle with semantic coherence, resulting in concept blending or attribute errors that fail to represent the input prompt accurately. This observation visually corroborates their lower scores on alignment benchmarks, such as T2I-CompBench. Furthermore, these methods often produce images plagued by perceptible artifacts, such as blurred textures and structural inconsistencies. These issues are symptomatic of the error accumulation inherent in their respective designs, such as the artifacts introduced by upsampling operators in spatial methods or the staleness of features in caching techniques.

In stark contrast, our JiT framework consistently produces high-fidelity, artifact-free images even at the demanding \(7\times\) speedup. A particularly challenging test is rendering legible text, a task that requires precise and high-frequency detail synthesis. As demonstrated, JiT successfully renders coherent text where other methods fail, producing results that are either garbled or incomplete. Collectively, these qualitative results provide compelling visual evidence that our JiT framework not only accelerates generation but also preserves the nuanced details and semantic integrity that substantially contribute to high-quality synthesis.

\vspace{-5px}

\begin{table}[t]
\renewcommand{\arraystretch}{0.8}
  \centering
  \begin{tabular}{lc}
    \toprule
    \textbf{Opponent Method} & \textbf{Preference for JiT (\%)} $\uparrow$ \\
    \midrule
    \multicolumn{2}{c}{\textit{JiT vs. \(\sim\)4\(\times\) Baselines}} \\
    \midrule
    FLUX.1-dev (12) & 85.6 \\
    Bottleneck (14) & 90.3 \\
    RALU (18)       & 73.7 \\
    Teacache (28)   & 69.1 \\
    \midrule
    \multicolumn{2}{c}{\textit{JiT vs. \(\sim\)7\(\times\) Baselines}} \\
    \midrule
    FLUX.1-dev (7)  & 93.1 \\
    RALU (10)       & 71.2 \\
    Taylorseer (28) & 89.5 \\
    Teacache (28)   & 75.4 \\
    \bottomrule
  \end{tabular}
  \caption{Human preference rates for our JiT method in blind pairwise comparisons.}
    \label{tab:user_study_preference}
\end{table}

\subsection{User study}
We conducted a user study to evaluate human perceptual preferences for the generated images directly. The study employed a blind pairwise comparison setup, where participants were asked to choose the superior image in terms of overall visual quality and prompt fidelity from a pair generated by our JiT and a competing baseline. A total of 1000 independent preference votes were collected from 20 participants evaluating 50 different prompts. The quantitative results, summarized in Tab.~\ref{tab:user_study_preference}, show that in comparisons against all baselines, images generated by our JiT framework were selected as the preferred option in a significant majority of votes, clearly demonstrating its leading position in human subjective evaluation.

%\vspace{-5px}
\subsection{Ablation study}
\begin{table}[t]
\centering
\setlength{\tabcolsep}{4pt}
\renewcommand{\arraystretch}{0.8}

\begin{tabular}{p{2.6cm}cc}
\toprule
Method & HPSv2.1 $\uparrow$ & T2I-Comp. $\uparrow$ \\
\midrule
\textbf{Ours} & \textbf{26.90} & \textbf{0.3727} \\
w/o SAG-ODE & 24.18 & 0.3414 \\
w/o ITA & 26.51 & 0.3670 \\
w/o DMF target & 26.04 & 0.3602 \\
\bottomrule
\end{tabular}

\caption{Ablation study of each component with JiT.}
\label{tab:ab}
\end{table}

\begin{figure}[b]
  \centering

   \includegraphics[width=\linewidth]{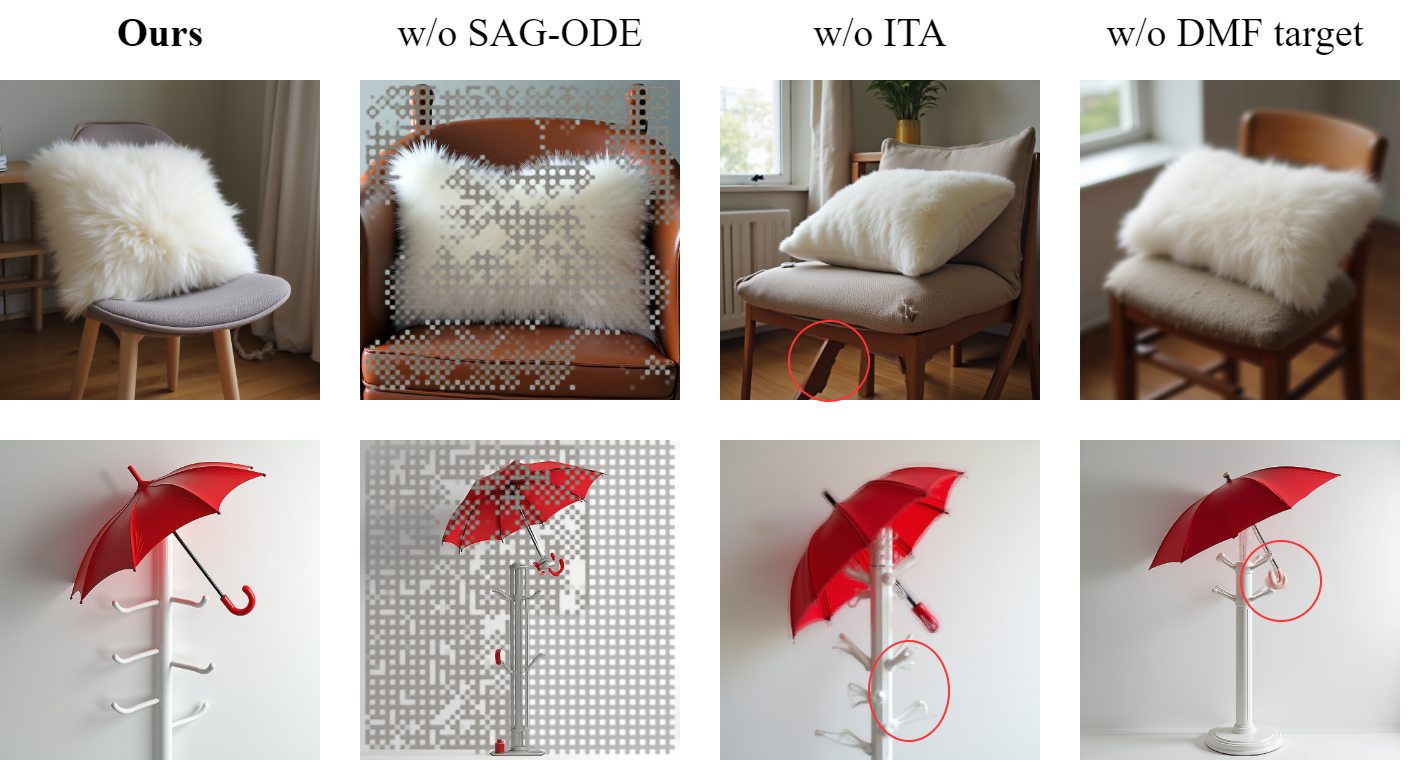}

   \caption{Visual ablation study of each component within JiT.}
   \label{fig:ab}
\end{figure}
In this section, we conduct a series of ablation studies to validate the effectiveness of each prominent component within our JiT framework. For computational efficiency, our ablation experiments are evaluated exclusively on the complex compositions subset of the T2I-CompBench \cite{t2icompbench} benchmark. Quantitative results are summarized in Tab.~\ref{tab:ab}, with corresponding qualitative comparisons provided in Fig.~\ref{fig:ab}. Further experiments on the schedule of JiT can be found in Appendix C.
%\vspace{-5px}
\paragraph{Effect of spatial approximation}
First, we ablate the core of our SAG-ODE by removing the spatial approximation term \(\mathcal{I}_k(\bm{u}_{\bm{\theta}})\) from our augmented lifter operator (Eq.~\eqref{eq:lifter}). This variant involves setting the velocity of all inactive tokens to zero, allowing them to evolve passively. As seen in Tab.~\ref{tab:ab}, this results in a catastrophic drop in performance. The qualitative results reveal the underlying cause: without the structurally aware velocity provided by our approximation, the inactive regions fail to form coherent structures and degenerate into meaningless noise, demonstrating that sophisticated extrapolation is essential.
%\vspace{-5px}
\paragraph{Effect of ITA}
Next, we replace our dynamic, importance-guided token activation strategy with a static, predefined selection pattern (e.g., a fixed grid). The visual examples in Fig.~\ref{fig:ab} demonstrate that a static pattern can misallocate computational resources. This leads to the emergence of artifacts and a loss of sharpness in complex, high-frequency regions that are not prioritized by the fixed schedule, highlighting the benefits of our content-aware approach.
%\vspace{-5px}
\paragraph{Effect of the DMF target}
Finally, we investigate the impact of our target state construction within the DMF. We replace our target Eq.~\eqref{eq-target-state} with a naive interpolation of the existing anchor tokens, which neglects the correct noise magnitude required by the flow trajectory. This modification results in a decline in performance, particularly in terms of perceptual quality. The qualitative results indicate that an improperly constructed target results in a noise mismatch during the transition.

\vspace{-7px}
\section{Conclusions}
We have introduced JiT, a novel and training-free framework that exploits spatial redundancy in DiT for significant inference acceleration. By formulating the SAG-ODE driven by a dynamic subset of tokens and ensuring seamless state transitions with a DMF, our method achieves a superior speed-fidelity trade-off. Our extensive experiments demonstrate up to \(7\times\) acceleration on the FLUX.1-dev model with negligible quality degradation, outperforming existing acceleration methods. By moving away from uniform computation towards a dynamic, on-demand strategy, this work not only provides a practical solution for accelerating generative ODEs but also opens new avenues for efficient, high-fidelity generative modeling.

\clearpage

{
    \small
    \bibliographystyle{ieeenat_fullname}
    \bibliography{main}
}

\appendix

 \clearpage
\setcounter{page}{1}
%\maketitlesupplementary
\onecolumn

\centering
\Large
\textbf{\thetitle}\\
\vspace{0.5em}Supplementary Material \\
\vspace{1.0em}
\raggedright
\normalsize

\section{The detailed derivation of SAG-ODE}
\label{A}
This section provides the formal derivation for the SAG-ODE introduced in the main paper. We restate the governing equation for a given stage $k$, which operates on the active token set $\Omega_{k}$:
\begin{equation}
\frac{\mathrm{d}\mathbf{y}(t)}{\mathrm{d}t} = \mathbf{\Pi}_{k}\, \bm{u}_{\bm{\theta}}(\mathbf{S}^\top_{k} \mathbf{y}(t), t).
\end{equation}
Here, $\mathbf{\Pi}_{k}$ is the augmented lifter operator, and the term $\bm{u}_{\bm{\theta}}(\mathbf{S}_{k}^\top \mathbf{y}(t), t)$ represents the velocity computed by the Transformer using only the $m_{k}$ active tokens. The following derivation shows how this compact form is obtained by starting from an exact orthogonal decomposition of the ideal velocity field and then applying the proposed approximation.

\subsection{Derivation via orthogonal decomposition}
Let $\bm{v}^*(\mathbf{y}, t) = \bm{u}_{\bm{\theta}}(\mathbf{y}, t) \in \mathbb{R}^{Nd}$ denote the ideal, full-dimensional velocity field that acts on all $N$ tokens. Using the projection matrix $\mathbf{P}_k = \mathbf{S}_k\mathbf{S}^\top_k$, we can decompose this velocity field into two orthogonal components: One lying within the subspace spanned by the anchor tokens (active subspace) and the other in its orthogonal complement (inactive subspace). The decomposition is given by
\begin{equation}
\bm{v}^* = \mathbf{P}_k \bm{v}^* + (\mathbf{I} - \mathbf{P}_k) \bm{v}^*.
\label{eq:ortho_decomp}
\end{equation}
Substituting $\mathbf{P}_k = \mathbf{S}_k\mathbf{S}^\top_k$ into the first term, we obtain
\begin{equation}
\bm{v}^* = \mathbf{S}_k (\mathbf{S}^\top_k \bm{v}^*) + (\mathbf{I} - \mathbf{P}_k) \bm{v}^*,
\end{equation}
where $\mathbf{S}^\top_k \bm{v}^* \in \mathbb{R}^{m_k d}$ represents the precise velocity projected onto the $m_k$ anchor tokens.

\subsection{Approximation strategy}
Computing the full velocity $\bm{v}^*$ is computationally prohibitive. To accelerate inference, we introduce two coherent approximations corresponding to the two terms in the decomposition:

\begin{enumerate}
    \item \textbf{Sparse input approximation (active component):} Instead of computing the velocity based on the full context $\mathbf{y}$, we approximate the active component by evaluating the network only on the projected anchor tokens $\mathbf{S}^\top_k \mathbf{y} \in \mathbb{R}^{m_k d}$. We define this computable sparse velocity as
    \begin{equation}
    \tilde{\bm{u}} := \bm{u}_{\bm{\theta}}(\mathbf{S}^\top_k \mathbf{y}, t) \in \mathbb{R}^{m_k d}.
    \end{equation}
    Thus, the first term in Eq.~\eqref{eq:ortho_decomp} is approximated by the embedding of this sparse calculation $\mathbf{S}_k \tilde{\bm{u}} \in \mathbb{R}^{N d}$.

    \item \textbf{Subspace extrapolation (inactive component):} For the second term $(\mathbf{I} - \mathbf{P}_k) \bm{v}^*$, which represents the velocity of the inactive tokens, we avoid explicit neural function evaluations. Instead, we assume that the dynamics of the inactive tokens can be reasonably inferred from the active anchors via a pre-defined interpolation operator $\mathcal{I}_k$ (please refer to Appendix \ref{inter} for specific implementation details). This yields the approximation
    \begin{equation}
    (\mathbf{I} - \mathbf{P}_k) \bm{v}^* \approx \mathcal{I}_k(\tilde{\bm{u}}).
    \end{equation}
\end{enumerate}

\subsection{Synthesis and the augmented lifter}
Substituting the approximations from the previous section, the approximate velocity field \(\bm{v}\) is formed by combining the exact velocity for the anchor tokens with the interpolated velocity for the inactive ones. We now define the augmented lifter \(\mathbf{\Pi}_k\) as the operator that encapsulates this entire process. Its action on the computed sparse velocity \(\tilde{\bm{u}}\) is given by
\begin{equation}
\bm{v} = \mathbf{\Pi}_k\tilde{\bm{u}} := \mathbf{S}_k \tilde{\bm{u}} + \mathcal{I}_k(\tilde{\bm{u}}).
\label{eq:lifter_action}
\end{equation}
This definition states that the operator \(\mathbf{\Pi}_k\) maps the sparse velocity to the full space. By substituting \(\tilde{\bm{u}} = \bm{u}_{\bm{\theta}}(\mathbf{S}^\top_k \mathbf{y}, t)\), we arrive at the final, compact form of our SAG-ODE:
\begin{equation}
\frac{\mathrm{d}\mathbf{y}(t)}{\mathrm{d}t} = \mathbf{\Pi}_{k}\bm{u}_{\bm{\theta}}(\mathbf{S}^\top_{k} \mathbf{y}(t), t).
\end{equation}

\subsection{Proof of consistency}
 Applying the projection $\mathbf{S}^\top_k$ to the derived ODE confirms that the dynamics on the activate anchor tokens remain exact relative to the sparse computation:
\begin{align}
\mathbf{S}^\top_k \frac{\mathrm{d}\mathbf{y}}{\mathrm{d}t} &= \mathbf{S}^\top_k (\mathbf{S}_k \tilde{\bm{u}} + \mathcal{I}_k(\tilde{\bm{u}})) \\
&= (\mathbf{S}^\top_k \mathbf{S}_k) \tilde{\bm{u}} + \mathbf{S}^\top_k \mathcal{I}_k(\tilde{\bm{u}}) \\
&= \mathbf{I} \cdot \tilde{\bm{u}} + \mathbf{0} \\
&= \bm{u}_{\bm{\theta}}(\mathbf{S}^\top_k \mathbf{y}, t).
\end{align}

Here, we utilize the property that $\mathbf{S}^\top_k \mathbf{S}_k = \mathbf{I}$ (orthonormality of the selector) and $\mathbf{S}^\top_k \mathcal{I}_k(\cdot) = \mathbf{0}$ (the interpolation operator has no effect on the anchor subspace by design). This concludes the derivation.

\section{Implementation details}
\label{sec:appendix_b}

In this section, we provide further implementation details of our JiT framework, including the specific construction of the selector matrix \(\mathbf{S}_k\) for the initial stage, the algorithm for our interpolation operators \(\mathcal{I}_k\) and \(\Phi_k\), and the detailed hyperparameter configurations for the experimental results reported in the main text.

\begin{figure*}[ht]
  \centering
  \includegraphics[width=0.6\linewidth]{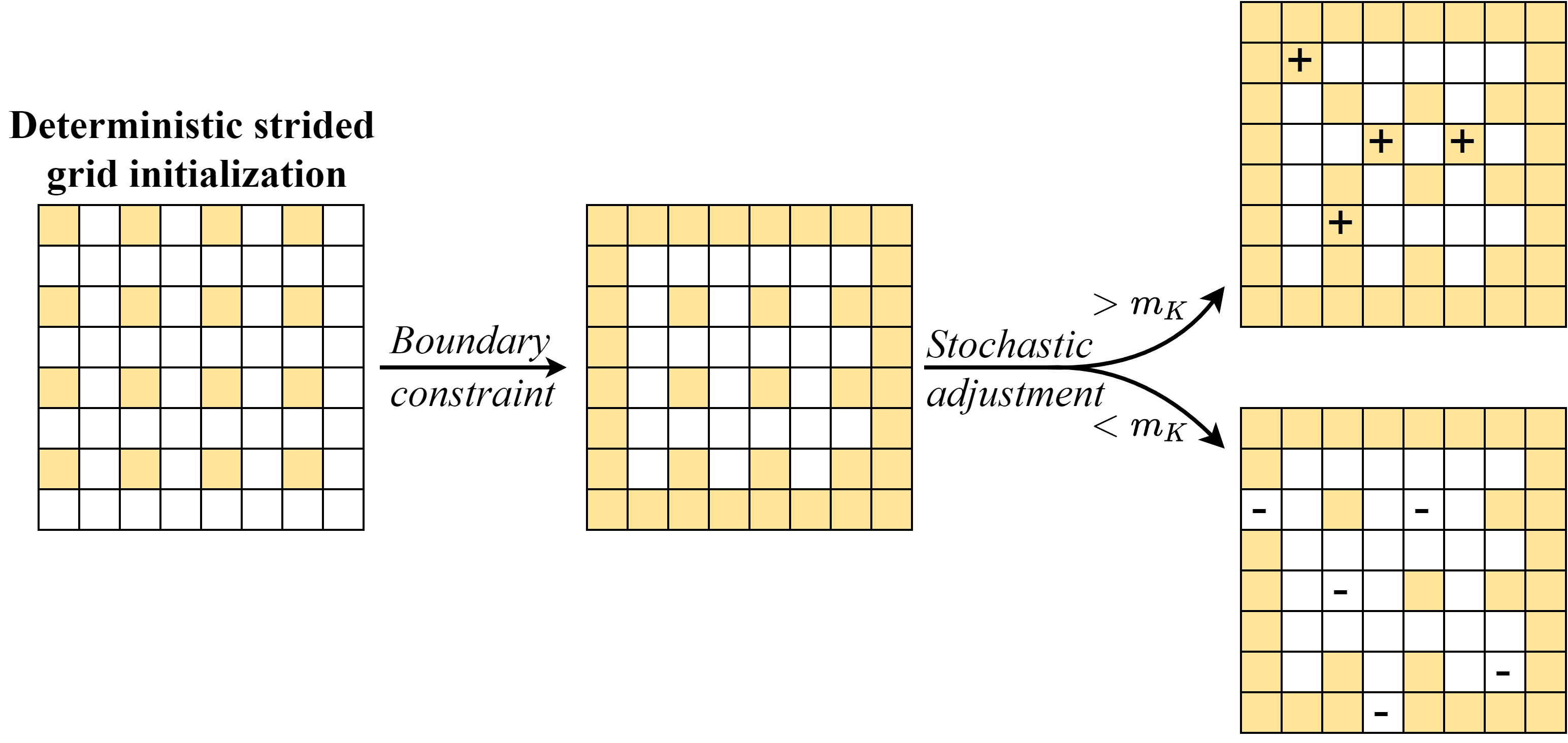}
  \caption{An illustration of the construction process for the initial selector matrix \(\mathbf{S}_K\). }
  \label{fig:s_k_selection}
\end{figure*}

\subsection{Construction of the initial selector matrix}
As described in Sec.~\ref{sec:token_selection}, for all subsequent stages \(k < K\), anchor tokens are selected dynamically based on the importance map derived from the velocity field variance. Consequently, we only need to define the construction of the selector matrix \(\mathbf{S}_K\) for the initial and most sparse stage.

For this initialization, our primary objective is to capture the global semantic structure while strictly adhering to the sparsity budget. To achieve this, we employ a deterministic strided grid initialization strategy, reinforced by explicit boundary constraints. Specifically, this strategy subsamples the latent grid with a fixed stride of 2 in both the height and width dimensions, which is akin to selecting the top-left token from every non-overlapping \(2 \times 2\) block. To ensure the integrity of the image frame and prevent boundary artifacts, we augment this set by explicitly including all tokens located at the top, bottom, left, and right boundaries of the latent feature map.

Since the number of tokens resulting from this geometric pattern may not strictly align with the target budget \(m_K\), we apply a final stochastic adjustment step. If the number of geometrically selected tokens falls short of \(m_K\), we randomly sample supplementary indices from the remaining inactive set to fill the deficit. Conversely, if the count exceeds \(m_K\), we randomly drop excess tokens from the generated set to strictly meet the \(m_K\) budget. This hybrid approach combines the regularity required for low-frequency structure generation with the flexibility to meet precise computational constraints. We provide a simplified schematic diagram of the above process in \cref{fig:s_k_selection}.

\subsection{Implementation of interpolation operators}
\label{inter}
It should be clarified that while sharing the same algorithmic implementation, the two interpolation operators serve distinct roles within our framework, the subspace extrapolation operator \(\mathcal{I}_k\) acts on the velocity field \(\bm{u}_\theta\), whereas the structural prior operator \(\Phi_k\) operates on the data latent tokens $\hat{\mathbf{y}}$.

We opt for a combination of nearest-neighbor interpolation and masked Gaussian blur rather than standard linear (e.g., bilinear) interpolation. This choice is driven by the nature of the high-dimensional representations in DiTs. Velocity fields and latent states often exhibit high-frequency spatial variations and complex semantic structures. Linear-based interpolation assumes a smooth gradient between sparse anchors, which can lead to over-smoothed features that drift off the valid feature manifold, particularly when the spatial grid is irregular or highly sparse. 

In contrast, nearest-neighbor interpolation strictly preserves the original feature statistics and signal integrity of the anchor tokens. We then address the resulting spatial discontinuities (blockiness) via a controlled Gaussian blur, which smooths the transition boundaries without corrupting the feature values at the anchor positions. Specifically, derived from the token sparsity $\rho$, we approximate the average anchor gap as $L \approx \rho^{-1/2}$. We dynamically set the blur scale $\sigma = 0.4L$ such that the effective radius of the Gaussian kernel ($3\sigma \approx 1.2L$) slightly exceeds and covers this gap. This geometric constraint systematically suppresses blocky artifacts while strictly preventing over-smoothing.

We present the unified implementation in Algorithm~\ref{alg:interpolation}.

\begin{algorithm*}[h]
\caption{Unified Interpolation Operator ($\mathcal{I}_k$ and $\Phi_k$)}
\label{alg:interpolation}
\begin{algorithmic}[1]
\STATE \textbf{Input:}Number of active tokens $m$, total tokens $N$, token dimension $d$, active features $\mathbf{Z}_\mathbf{act} \in \{\bm{u}_\mathbf{act}, \hat{\mathbf{y}}_\mathbf{act}\} \in \mathbb{R}^{md}$, active indices \(\mathcal{A}_\mathbf{act}\) 

\STATE \textbf{Output:} Interpolated full tensor $\mathbf{Z}_{\mathbf{full}} \in \{\bm{u}_{\mathbf{full}}, \hat{\mathbf{y}}_{\mathbf{full}}\} \in \mathbb{R}^{Nd}$

\STATE \textbf{Step 1: Nearest-Neighbor Interpolation}
\STATE Generate full coordinate grid \(\mathbf{C}_{\mathbf{full}}\) and active coordinates \(\mathbf{C}_\mathbf{act}\)
\FOR{each token $i \in \{1, \dots, N\}$}
    \STATE $j^* = \arg\min_{j \in \mathcal{A}_\mathbf{act}} \|\mathbf{C}_{\mathbf{full}}[i] - \mathbf{C}_\mathbf{act}[j]\|_2$
    \STATE $\mathbf{Z}_{\mathbf{NN}}[i] \leftarrow \mathbf{Z}_\mathbf{act}[j^*]$
\ENDFOR

\STATE \textbf{Step 2: Controlled Gaussian Blur}
\STATE Calculate token density $\rho = m / N$ and approximate anchor gap $L \approx \rho^{-1/2}$
\STATE Set blur scale $\sigma = 0.4L$ and kernel size $k \approx 3\sigma$ \COMMENT{Ensure $k > L$ for adequate spatial coverage}
\STATE $\mathbf{Z}_{\mathbf{blur}} \leftarrow \text{GaussianBlur}(\mathbf{Z}_{\mathbf{NN}}, \text{kernel}=k, \sigma=\sigma)$

\STATE \textbf{Step 3: Masked Composition}
\STATE Initialize binary mask $\mathbf{M} \leftarrow \mathbf{0}_N$
\STATE $\mathbf{M}[\mathcal{A}_\mathbf{act}] \leftarrow 1$ \COMMENT{Identify anchor positions}
\STATE $\mathbf{Z}_{\mathbf{full}} \leftarrow \mathbf{M} \odot \mathbf{Z}_{\mathbf{NN}} + (\mathbf{1} - \mathbf{M}) \odot \mathbf{Z}_{\mathbf{blur}}$ \COMMENT{Preserve exact anchor values}

\RETURN $\mathbf{Z}_{\mathbf{full}}$
\end{algorithmic}
\end{algorithm*}

\subsection{Detailed experimental configurations}
As shown in \cref{tab:jit_settings}, we provide the detailed hyperparameter settings for our JiT framework under the $4\times$ and $7\times$ acceleration settings reported in the main paper. The configuration includes the total NFEs, the stage definition, and the specific token sparsity ratios used at each stage.

It is important to note that these configurations are not the product of an exhaustive, brute-force search. Instead, they are derived from a principled and lightweight hyperparameter recipe. Specifically, we first determine an initial token sparsity threshold that strictly ensures semantic stability at the generation starting point. To achieve the target acceleration ratios (e.g., $4\times$ or $7\times$), we then strategically allocate the solver steps, assigning fewer steps to the highly sparse early stages to maximize inference speed, and dedicating more steps to the denser late stages to preserve fine-grained generation quality.

\begin{table}[ht]
\renewcommand{\arraystretch}{0.9}
\centering
\begin{tabular}{l|cc}
\toprule
\textbf{Parameter} & \textbf{JiT (\(\sim 4\times\))} & \textbf{JiT (\(\sim 7\times\))} \\
\midrule
Total NFEs & 18 & 11 \\
Number of Stages & 3 & 3 \\
\midrule
\textbf{Stage 2 (Coarse)} & & \\
\quad Steps & 7 & 4 \\
\quad Token Sparsity & 35\% & 32\% \\
\midrule
\textbf{Stage 1 (Medium)} & & \\
\quad Steps & 4 & 3 \\
\quad Token Sparsity & 62\% & 60\% \\
\midrule
\textbf{Stage 0 (Fine)} & & \\
\quad Steps & 7 & 4 \\
\quad Token Sparsity & 100\% & 100\% \\
\bottomrule
\end{tabular}
\caption{Detailed configuration of JiT schedules for different acceleration factors. The sparsity ratio denotes the fraction of tokens activated at each stage relative to the full sequence length \(N\).}
\label{tab:jit_settings}
\end{table}

\subsection{Timestep schedule configuration}
\label{sec:beta_schedule}

Standard uniform timestep sampling typically assumes a constant rate of information generation, which fails to capture the dynamic nature of the diffusion process. Recent spectral analysis in Beta Sampling~\cite{beta} reveals that significant changes in image content, specifically the establishment of low-frequency global structures, occur predominantly in the early stages of the denoising process, while high-frequency details are refined later.

To align the computational effort with this non-uniform information flow, we adopt a Beta-distribution-based strategy to define our timestep schedule. The core methodology leverages the probability integral transform (PIT) to warp the standard uniform timestep discretization. Formally, given a linear sequence \(s_i \in [0, 1]\) representing the uniform progress of generation, we generate a non-uniform set of timesteps by mapping them through the inverse cumulative distribution function (CDF) of the Beta distribution. The final timesteps \(t_i\) corresponding to these warped time points are then calculated as
\begin{equation}
t_i = F^{-1}(s_i; \alpha, \beta),
\end{equation}
where \(F^{-1}(\cdot; \alpha, \beta)\) denotes the inverse CDF of the Beta distribution. This procedure effectively creates a non-uniform sequence of timesteps \(\{t_i\}\) for the ODE solver.

In the context of our JiT framework, the robust formation of the global structure in the early phase is a prerequisite for the effectiveness of the subsequent spatial acceleration. Therefore, we prioritize the early denoising stages by allocating a denser distribution of timesteps to this period. We implement this by setting the hyperparameters to \(\alpha=1.4\) and \(\beta=0.42\). This configuration skews the sampling density towards the high-noise region (i.e., early in the denoising process). This ensures that the model dedicates sufficient NFEs to solidly establish the semantic layout and structural foundation before transitioning to the accelerated, spatially-sparse refinement stages.

\subsection{Evaluation dataset}
\label{sec:eval_dataset}

To ensure a comprehensive and unbiased evaluation, we constructed our evaluation dataset using the official prompt sets from the GenEval~\cite{geneval} and T2I-CompBench~\cite{t2icompbench} benchmarks. 

Specifically, for the 553 prompts from GenEval, we generated four images per prompt, each with a different random seed, yielding a subset of 2,212 images (553 prompts \(\times\) 4 seeds). For the 2,400 prompts from T2I-CompBench, we generated one image per prompt. 

Therefore, our full evaluation dataset comprises a total of 4,612 images, providing a robust and diverse foundation for the metrics reported in our experiments.

\section{Additional ablation study on JiT scheduling}
\label{sec:appendix_c}

In this section, we provide further qualitative ablation studies on the scheduling hyperparameters of our JiT framework. These experiments are designed to offer insight into how different scheduling choices impact the trade-off between computational cost and generation quality, thereby justifying the configurations used in our main experiments.

\subsection{Impact of the number of stages}
The number of stages in the JiT schedule plays a significant role in defining the granularity of the coarse-to-fine generation process. To investigate this, we compare our default 3-stage setup against 2-stage and 4-stage variants, keeping the total NFEs fixed at 18. For fairness, the sparsity ratios and transition timings in these variants are set to be approximately uniform (e.g., 50\% sparsity for the first stage in the 2-stage schedule).

\begin{figure*}[ht]
  \centering
  \includegraphics[width=0.5\linewidth]{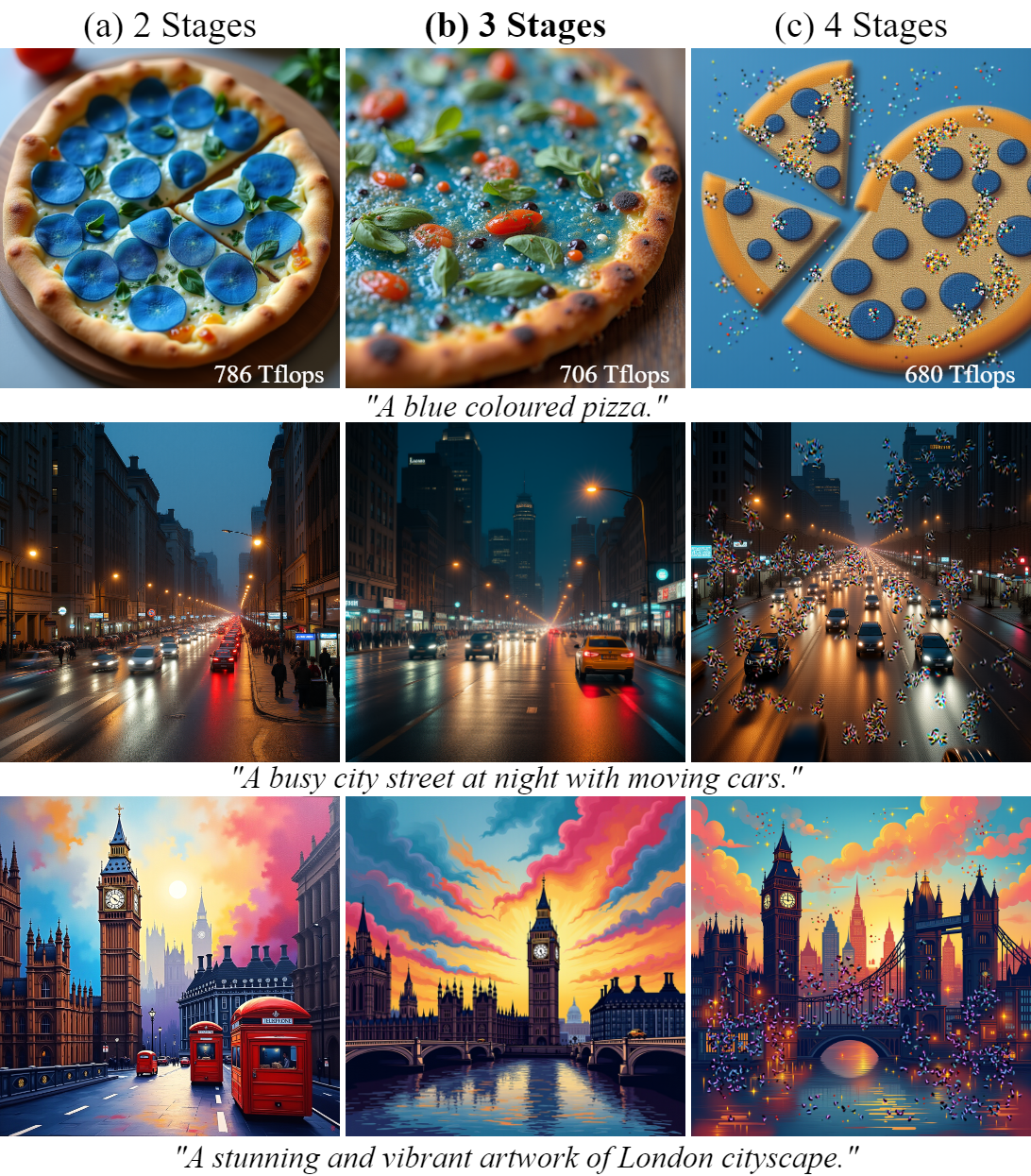}
  \caption{Visual ablation on the number of stages for a fixed total NFE of 18. (a) A 2-stage schedule offers limited acceleration. (b) Our default 3-stage schedule yields the best trade-off. (c) A 4-stage schedule introduces persistent noise due to a late transition to full resolution. This validates that a 3-stage approach offers a superior balance between speed and quality.}
  \label{fig:ablation_stages}
\end{figure*}

As visually demonstrated in Fig.~\ref{fig:ablation_stages}, our default 3-stage schedule strikes the most effective balance. 
(a) The 2-stage schedule (50\% \(\to\) 100\%) offers limited acceleration and produces a slightly blurred image, a quality degradation we attribute to the large, abrupt incorporation of half the tokens simultaneously.
(b) Our default 3-stage schedule avoids both pitfalls, providing substantial acceleration in the early phases while allowing sufficient time for detail refinement in the later, denser stages.
(c) Conversely, the 4-stage schedule, while computationally cheaper, suffers from noticeable quality degradation. Its final transition to full resolution occurs too late in the denoising process. By this point, the global structure is largely solidified, and the late introduction of high-frequency information results in persistent noise and artifacts that the model cannot fully remove in the remaining few steps.

\subsection{Impact of token allocation (sparsity ratios)}
\begin{figure*}[ht]
  \centering
  \includegraphics[width=0.5\linewidth]{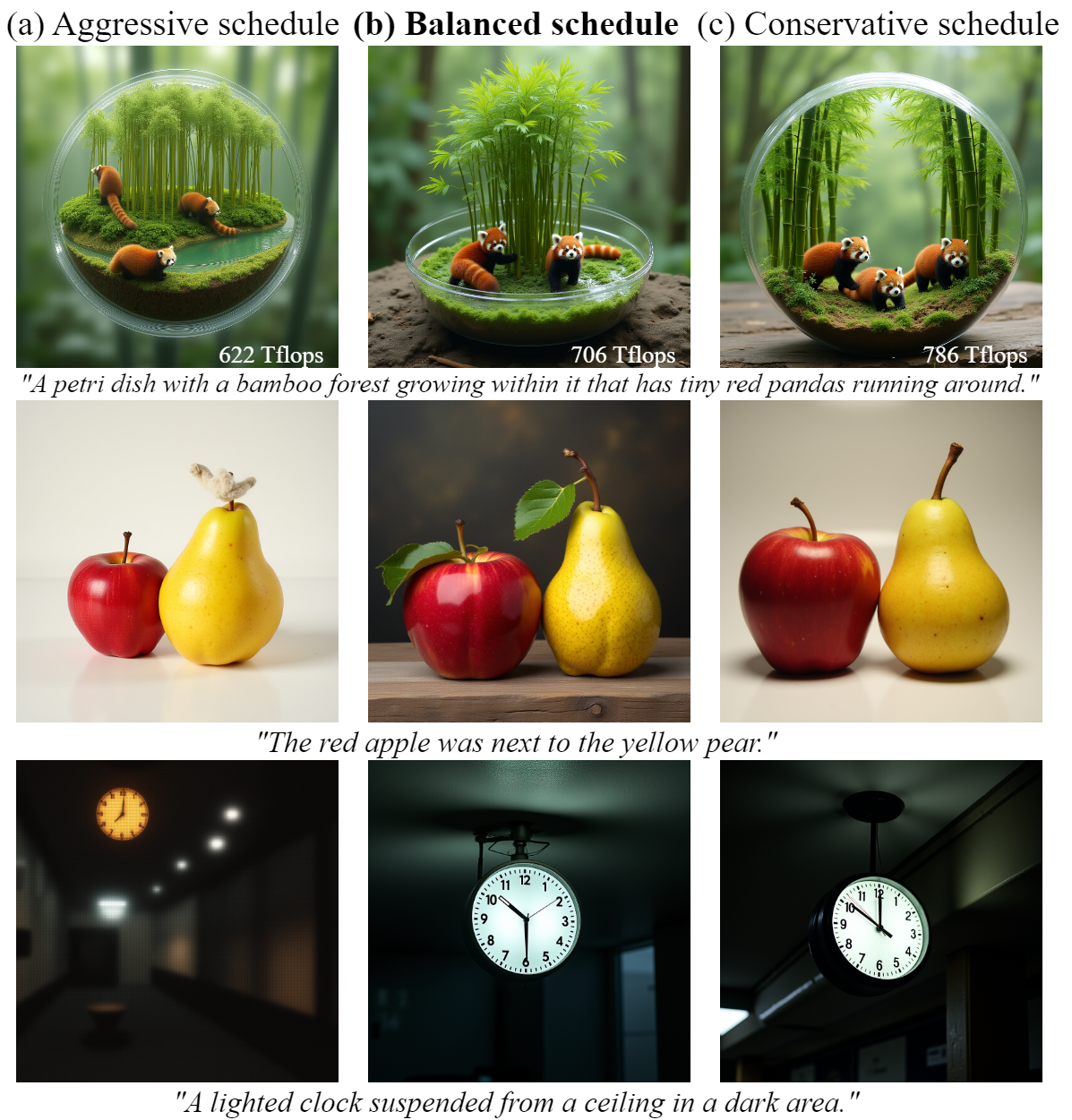}
  \caption{Visual ablation on token allocation for a fixed 3-stage, 18-NFE schedule. (a) An aggressive schedule (20\% \(\to\) 50\%) compromises semantic integrity. (b) Our chosen schedule (35\% \(\to\) 62\%) provides the best balance. (c) A conservative schedule (50\% \(\to\) 75\%) offers similar quality but with a reduced acceleration advantage.}
  \label{fig:ablation_allocation}
\end{figure*}

Given a fixed 3-stage schedule, the allocation of tokens to the early and middle stages directly controls the computational cost and the quality of the structural prior. A more aggressive schedule (fewer tokens) saves more computation but risks creating a poor initial structure that is difficult to correct later. We explore this trade-off in Fig.~\ref{fig:ablation_allocation} by comparing our default schedule (35\% \(\to\) 62\% \(\to\) 100\%) with two variants: An aggressive schedule (20\% \(\to\) 50\% \(\to\) 100\%) and a conservative one (50\% \(\to\) 75\% \(\to\) 100\%).

The results are revealing. (a) The aggressive schedule, despite its significant acceleration, yields an image with compromised semantic integrity and lower overall quality. This suggests that an overly sparse initial set of tokens fails to capture sufficient global information, leading to errors that propagate through the entire generation process. (b) In contrast, our chosen token allocation robustly preserves generation quality while still achieving substantial acceleration. (c) On the other hand, while the conservative schedule also produces a high-quality result comparable to our default, its acceleration advantage is substantially diminished due to the high token count in the initial stages. This three-way comparison validates that our chosen approach provides the most effective compromise.

\section{Extensibility to other backbone models}
\label{sec:appendix_d}
To demonstrate the generalizability of our JiT framework and its applicability beyond the FLUX.1-dev model, we conducted an additional experiment applying JiT to a different, publicly available backbone: Qwen-image~\cite{qwen}. Qwen-image is another powerful text-to-image model based on a DiT architecture, but with a distinct design and training paradigm compared to FLUX. This makes it an excellent test case for the model-agnostic nature of our approach.

\begin{figure*}[ht]
  \centering
  \includegraphics[width=\linewidth]{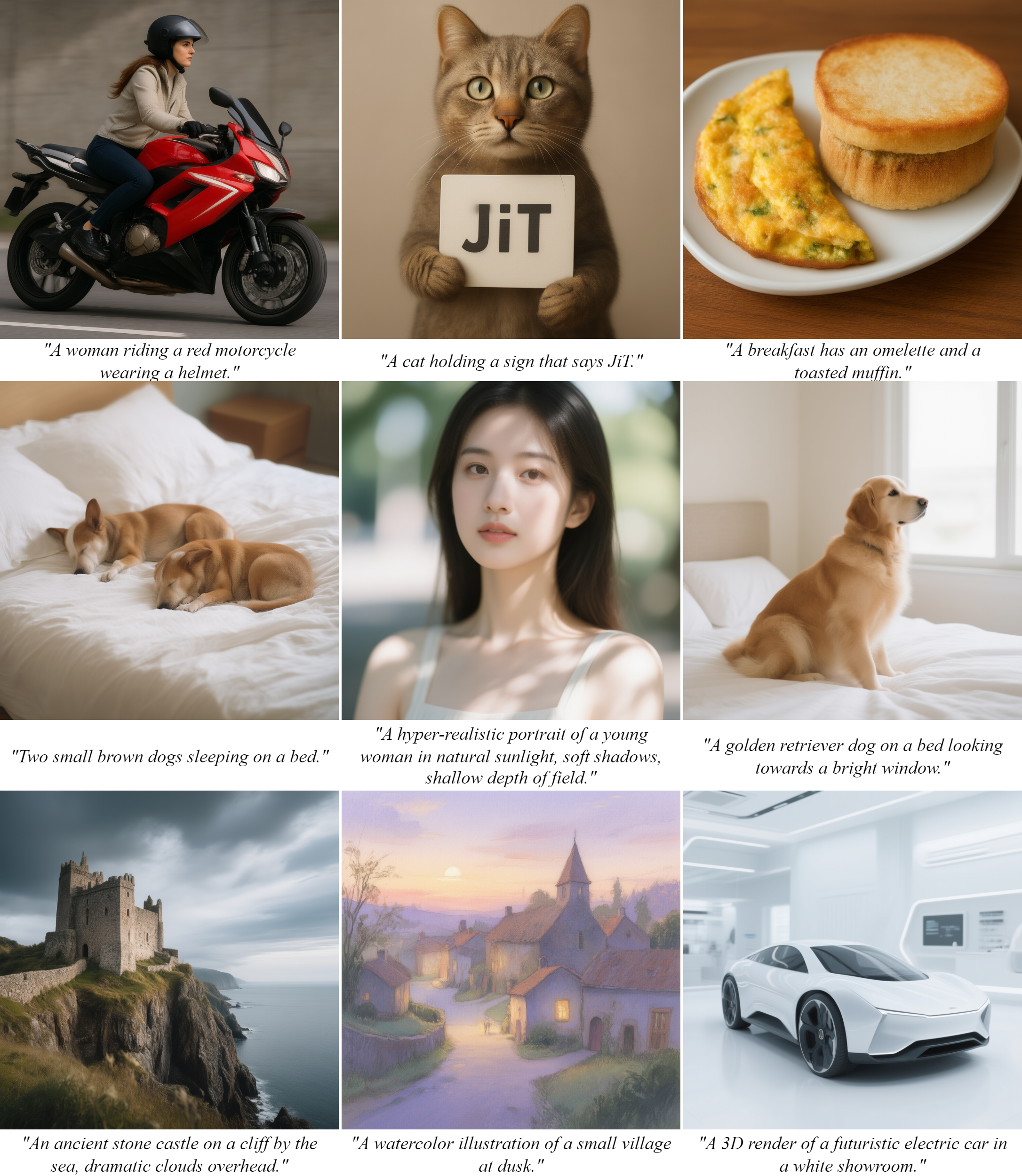}
  \caption{Qualitative results of our JiT framework applied to the Qwen-image model. The images were generated with $\sim$4$\times$ acceleration compared to the standard pipeline.}
  \label{fig:qwen_showcase}
\end{figure*}

By simply defining a multi-stage schedule with dynamic token selection and the corresponding SAG-ODE approximations, we are able to achieve a inference acceleration of approximately 4\(\times\) (from 26.95s to 6.51s). As shown in the visual results in \cref{fig:qwen_showcase}, the outcomes are highly compelling. Despite the substantial speedup, the model maintains a high degree of visual fidelity, successfully rendering complex scenes, textures, and semantic concepts presented in the prompts. 

Furthermore, to rigorously validate the scalability of our method across diverse modalities, we extend JiT to the spatiotemporal domain using the HunyuanVideo-1.5~\cite{video4} backbone. Video diffusion models inherently process 3D latent representations, exhibiting significant redundancy both spatially within a single frame and temporally across continuous frames. Because our spatial acceleration strategy treats latents as a unified sequence of discrete tokens, it seamlessly identifies and bypasses these redundant spatiotemporal computations without requiring architectural modifications.

\begin{figure*}[t]
    \centering
    \includegraphics[width=\linewidth]{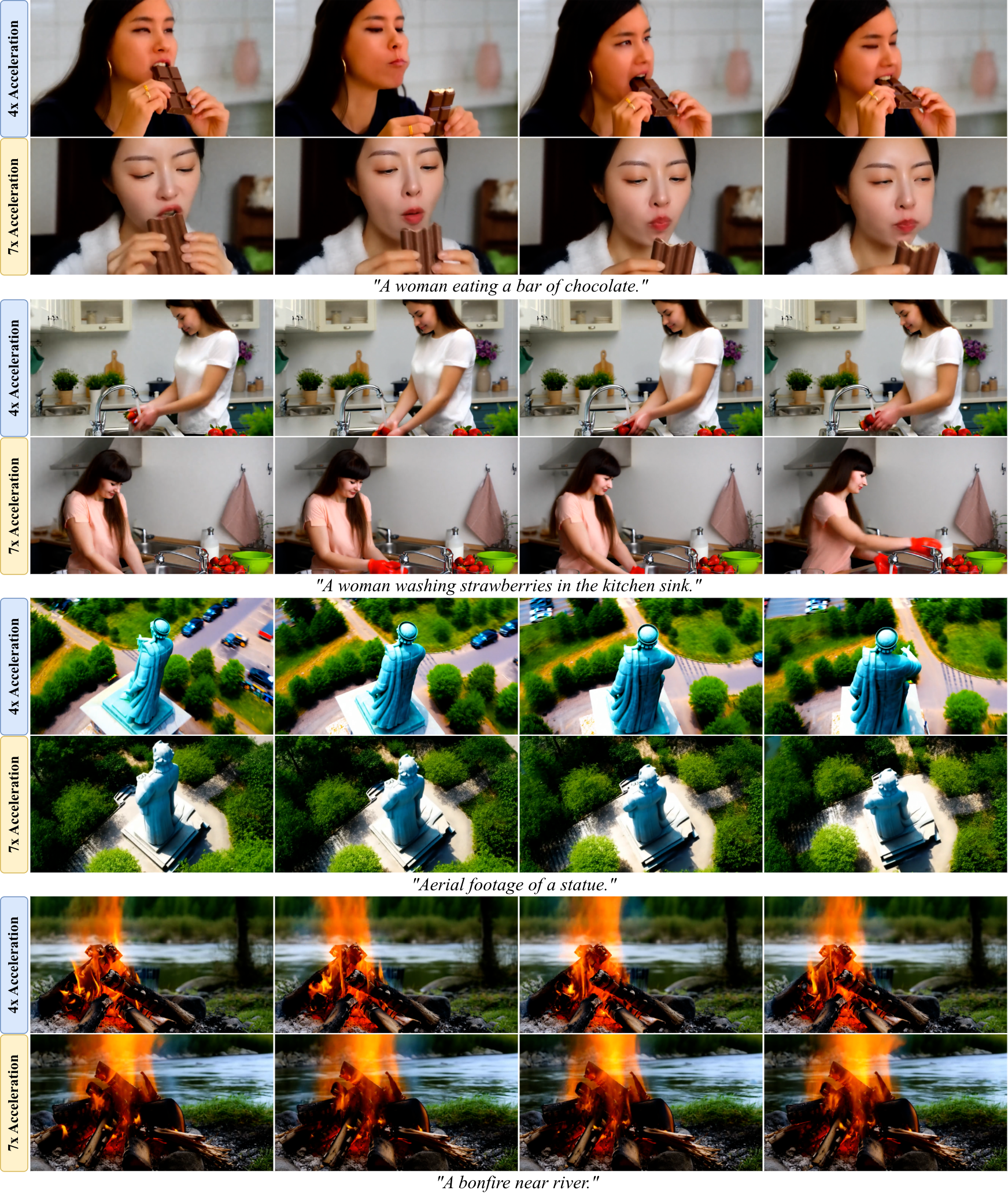} 
    \caption{Qualitative results of our JiT framework applied to the HunyuanVideo-1.5 backbone. Our JiT framework demonstrates robust generalizability in the spatiotemporal domain, maintaining semantic consistency and temporal coherence under both $\sim$4$\times$ and $\sim$7$\times$ acceleration settings.}
    \label{fig:hunyuan_showcase}
\end{figure*}

As illustrated in \cref{fig:hunyuan_showcase}, we provide qualitative showcases of the generated video frames under both $\sim 4\times$ (from 1830.21s to 423.52s) and $\sim 7\times$ (from 1830.21s to 268.12s) acceleration settings. The visual outcomes confirm that JiT effectively maintains motion coherence and spatial structural fidelity even at extreme speedup ratios. This successful cross-modality application serves as strong evidence that JiT relies on the fundamental coarse-to-fine nature of the diffusion probability flow, rather than over-fitting to the architectural biases of a specific 2D model.

\section{Additional qualitative comparisons}
\label{sec:appendix_e}

\begin{figure*}[ht]
  \centering
  \includegraphics[width=\textwidth]{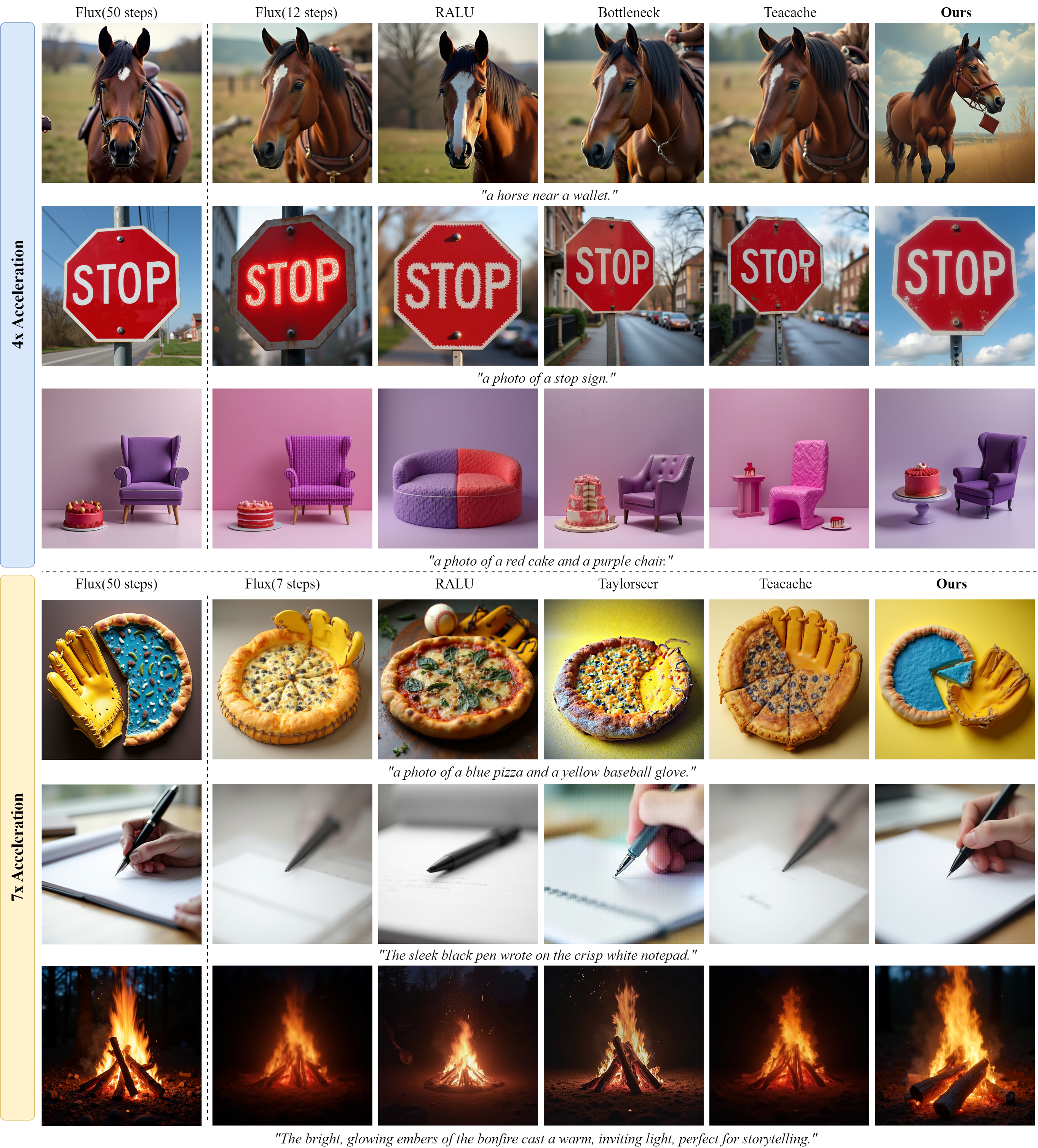}
  \caption{Additional qualitative comparisons of our JiT framework against baseline methods at \(\sim\)4\(\times\) and \(\sim\)7\(\times\) acceleration. These examples further showcase the superior performance of JiT across a variety of challenging prompts. 
  }
  \label{fig:additional_qualitative}
\end{figure*}

To further validate the robustness of our JiT framework, ~\cref{fig:additional_qualitative} presents additional qualitative comparisons on challenging prompts. The results consistently demonstrate the superiority of JiT over competing methods at similar computational budgets, particularly in preserving complex compositions and fine-grained details.

\end{document}